\pdfoutput=1

\documentclass[11pt]{article}

\usepackage[final]{acl}

\usepackage{times}
\usepackage[disable]{todonotes}
\usepackage{latexsym}

\usepackage[T1]{fontenc}

\usepackage[utf8]{inputenc}

\usepackage{microtype}

\usepackage{inconsolata}

\newcommand{\fig}[3][0.9]{
\begin{figure}
\centering
\includegraphics[width=#1\columnwidth]{figure/#2}
        \caption{#3}~\label{fig:#2}
\end{figure}}

\title{Large Language Models as Minecraft Agents}

\author{Chris Madge\\
  Queen Mary University of London \\
  \texttt{c.j.madge@qmul.ac.uk} \\\And
  Massimo Poesio\\
  Queen Mary University of London \\
  \texttt{m.poesio@qmul.ac.uk} \\}

\begin{document}
\maketitle
\begin{abstract}
In this work we examine the use of Large Language Models (LLMs) in the challenging setting of acting as a Minecraft agent. We apply and evaluate LLMs in the builder and architect settings, introduce clarification questions and examining the challenges and opportunities for improvement.  In addition, we present a platform for online interaction with the agents and an evaluation against previous works.
\end{abstract}

\section{Introduction}
\label{sec:introduction}

\todo[inline]{I rearranged a bit the order of the paragraphs and added some requests for references but left the content more or less the same, it's quite good!}

The development of conversational agents able to operate in virtual world environments  has  long been of interest in AI \cite{winograd1972understanding}.
In recent years, much of this research has focused on developing agents able to operate in game environments. 
Game environments
provide an ideal sandbox for studying task-oriented
conversational agents in games \cite{szlam2019build},
which has motivated the development of multiple platforms in which such research can be carried out  \cite{johnson2016malmo,urbanek2019light,callison2022dungeons}
\todo[inline]{Add here references to Malmo, Light, and the more recent work on a Dungeons and Dragons platform; CM: added refs}\cite{gray2019craftassist,ogawa2020gamification,kohn2020mc}, data gathering exercises \cite{narayan2019collaborative,jayannavar2020learning,mohanty2022collecting} and competitions \cite{iglucomp}.
In this work we study the development of conversational agents able to operate in one of the most popular of these virtual world settings, the Minecraft world, and to carry out the ``Collaborative Builder Task'' proposed in \cite{narayan2019collaborative}, in which an architect, who has a target structure to achieve, provides instructions to a builder, to complete that structure.

In previous research, deep neural methods were used to model both of these roles.  These models  used a CNN to model the world state, and an RNN to model the text input, with a similar decoder to produce the utterances \cite{narayan2019collaborative,jayannavar2020learning,kiseleva2023interactive,shi2022learning}. An alternative approach has been to separate the builder task into interpretation of natural language instructions, translating them into structured commands using a fine tuned T5 model \cite{t5}, then using a separate model to enact the required actions in the world \cite{kiseleva2023interactive}.

However, while it has been proposed using modern large language models (LLMs) such as Llama and GPT-4 for this purpose has been proposed 
\cite{hromei2023end} \todo[inline]{REFS; CM: moved ref}, to our knowledge this has yet to be explored.
LLMs are arguably one of the most interesting advancements in natural language technology, if not the field of machine learning, in recent years. Some studies suggest that beyond their conversational ability (natural language understanding and generation \cite{achiam2023gpt}), such models demonstrate the ability to solve novel tasks akin to an artificial general intelligence \cite{bubeck2023sparks} with great potential for serving as the foundation for interactive agents \cite{xi2023rise}.
There are also many applications of LLMs as autonomous agents or embodied AI in interactive environments, particularly in robotics systems \cite{wu2023tidybot,ahn2022can}
and multi-agent simulations with cooperation/interaction \cite{park2023generative,lin2023agentsims,chen2023agentverse},
industrial automation \cite{xia2023towards} and planning \cite{nottingham2023embodied,wang2023voyager,zhu2023ghost}.
However, we are not aware of studies using LLMs to implement Minecraft agents. 



Our first contribution are LLM-based builder and architect agents. 
We developed builder and architect with both open-source and closed-source LLMs. 
We evaluate the builder by comparing it to previous comparable agents. 
Our second contribution is giving to these agents the ability to ask   clarification questions 
\cite{skrynnik2022learning}. 
This is one ability of conversational agents  that remains relatively unexplored in this setting; to our knowledge, there is only a single model attempting this \cite{mehta2023improving}, using BART \cite{bart}, but none with  current LLMs. 
Our third contribution is a test environment, which we believe is highly suited to devising and testing different approaches to these problems in a simple task orientated setting. We deliberately present our environment as a web-based online platform to support future large scale experimentation with human participation.
Our implementation draws inspiration from the web-based Interactive Grounded Language Understanding (IGLU) data collection system \cite{mohanty2022collecting,mehta2023improving}.

\section{Related Work}
\label{sec:related}

\todo[inline]{discussion on other LLM agents \cite{wang2023survey}}
\todo[inline]{For arxiv it is fine not to have a discussion of other work on LLMs for agents in games, but we should have that in the final version}

\subsection{The Original Architect and Builder Models for the Minecraft Task}

Several models for both the \textit{architect} and \textit{builder} roles  have been proposed. 
The first \textit{architect} model \cite{narayan2019collaborative} uses a seq2seq RNN that aside from the usual sequence of tokens as an input, also takes the current and target world state. This is decoded to the Architect utterance.
    
Following on from the Architect Model,  the ``Builder Action Prediction'' model was created \cite{jayannavar2020learning}. This model took also combined a CNN with an RNN, although the decoder now predicted a series of actions the agent could follow.

\subsection{IGLU}

The Interactive Grounded Language Understanding shared tasks \citep{iglucomp,kiseleva2022iglu} were proposed to stimulate research on building interactive agents that learn to solve a task while provided with grounded natural language instructions in a collaborative environment. 
The more recent edition of the shared task \cite{kiseleva2022iglu} involved two subtasks: 
`Building Structures' (developing a conversational agent able to engage in the conversation), and `Asking Clarification Questions' (developing a model able to ask clarification questions at the appropriate point). 

\todo[inline]{Are the two following paras about the IGLU 2021 or IGLU 2022? And for which task are the baselines for? CM: added clarification and citations}

The baseline \cite{skrynnik2022learning} provided for the 2022 IGLU task \cite{kiseleva2023interactive} is split into two separate modules. One language component responsible for interpreting the natural language utterances and transforming them into simplified structured instructions, and a reinforcement learning component responsible for navigating the agent in the world and performing actions to realise those instructions.  

    Several builder models were entered in the IGLU 2022 competition task
    \cite{kiseleva2022iglu}.  The creators of the competition reflect upon the most
    notable winning submissions \cite{kiseleva2023interactive}. The top
    architecture took an end-to-end approach concatenating outputs from a
    language model and a CNN, which is then fed into a transformer designed
    specifically to improve and fine tune language models through reinforcement
    learning. However, second and third place both made improvements on the
    baselines architecture approach with further fine tuning on the T5 transformer model \cite{t5}.

\subsection{Using LLMs to generate Embodied AI in an Interactive Environment}
There are many applications of LLMs as autonomous agents or embodied AI in interactive environments \cite{wang2023survey,xi2023rise}. Many of these do operate in a Minecraft world.  However, they are targeted towards planning (e.g. dividing tasks to reach a final goal, including ``crafting''/combining items, to reach that outcome) \cite{nottingham2023embodied,wang2023voyager,zhu2023ghost}, not the creation of 3D structures. 

\subsection{Generating and Responding to Clarification Questions}

Generating and responding to clarification questions generative remains relatively unexplored for this task. There have been models that label utterances as requiring clarification \cite{shi2022learning}, or rank clarification questions appropriateness to an instruction in an information retrieval style 
 \cite{kiseleva2023interactive,mohanty2023transforming}, but only one generative model \cite{mehta2023improving} using BART \cite{bart}. Whilst the notion of using modern LLMs such as Llama and GPT-4 for this purpose has been proposed, to our knowledge this is yet to be explored \cite{hromei2023end}. 



\section{Methodology}

In this work we created models for  the ``Collaborative Builder Task'' \cite{narayan2019collaborative} in which two agents collaborate in a voxel based world to build a structure.  The \textit{architect} is given a target world structure, and provides natural language instructions to a \textit{builder}, who cannot see the target structure, to place blocks to achieve that goal structure. Unlike the majority of previous approaches, both agents are able to provide and answer questions.

\subsection{The Builder}

The builder agent operates as follows:
The initial prompt to the LLM contains instructions for the task, constraints of the world ($min/max$ for $x,y,z$), coordinate mappings (east/south/west/north) and requests that the LLM respond in a specified JSON format only (see \ref{sec:builderprompt} for full prompt).  Following the initial prompt, on each turn, the agent consumes an utterance (instruction) and produces a sequence of actions in addition to an optional clarification question in the originally specified format.  This JSON encoded representation is used to directly update the world. If the result does not constitute a valid JSON response it is disregarded.
The builder is evaluated against IGLU RL baseline sub-task. We evaluate the agent based on its ability to interpret a natural language instruction and produce the correct 3D coordinates and colours for the target structure (not manoeuvre an agent in a 3D space to interact with the world and realise that target structure).

The closest work to our methodology is \cite{mehta2023improving}, similarly we focus solely on the language component of this task. We do not examine the task of manipulating an agent to place blocks as done in previous tasks \cite{narayan2019collaborative,jayannavar2020learning,shi2022learning}, as it is the language element that we are most interested in.

\subsection{The Architect}

The architect differs slightly from the builder.
The prompt, along with architect specific instructions, contains the target state of the world (see \ref{sec:architectprompt}). Following the initial prompt the architect consumes every utterance given by the builder, along with the current state of the world encoded in a structured JSON format.  The target structures are taken from the 
Minecraft Dialogue Corpus 
\cite{narayan2019collaborative} and evaluated empirically through repeated interactions.


\subsection{The LLMs}

In this work we test against the LLMs: GPT-4; GPT-3.5; Llama2 \cite{touvron2023llama} (the chat variation in various parameter sizes) and Vicuna \cite{vicuna2023}.  To our knowledge, the number of parameters used in the GPT models have not been formally released, although they are rumoured to be magnitudes greater than those used in the open models. The prompts used for both tasks are given in the appendix.

\todo[inline]{We need to provide details here regarding which LLMs we used, the prompts, etc (referring to the Appendix for examples of prompts); CM: added discussion and ref to appendix}

\section{Results}
\subsection{Builder}

Here, the builder is the agent, and the architect is either a human tester, or a dataset for evaluation purposes.

Table \ref{tbl:results} shows the results comparing the original IGLU Baseline against our LLM based builder agent.  GPT-4 and GPT-3.5 perform similarly, outperforming the IGLU NLP evaluation baseline \footnote{\url{https://gitlab.aicrowd.com/aicrowd/challenges/iglu-challenge-2022/iglu-2022-rl-mhb-baseline/-/blob/master/local_nlp_evaluation.py}}. However Llama2, and other LLM models do not.  Llama2-70b-chat comes close, but the smaller parameter models struggle to consistently adhere to the structured response format. This is particularly in evident in the case of Vicuna-13b which often generates responses in encoded in other languages/formats.

\begin{table}[ht!]
\centering
\begin{tabular}{r|l}
Model & Accuracy \% \\
\hline
IGLU Baseline & 27.1 \\
GPT-4 & \textbf{37.6} \\
GPT-3.5 & 34.4 \\
Llama2-70b-chat & 23.9 \\
Llama2-13b-chat & 15.0 \\
Vicuna-13b & 6.3 \\
\end{tabular}
\caption{Accuracy on IGLU NLP Evaluation }
\label{tbl:results}
\end{table}

\fig[1]{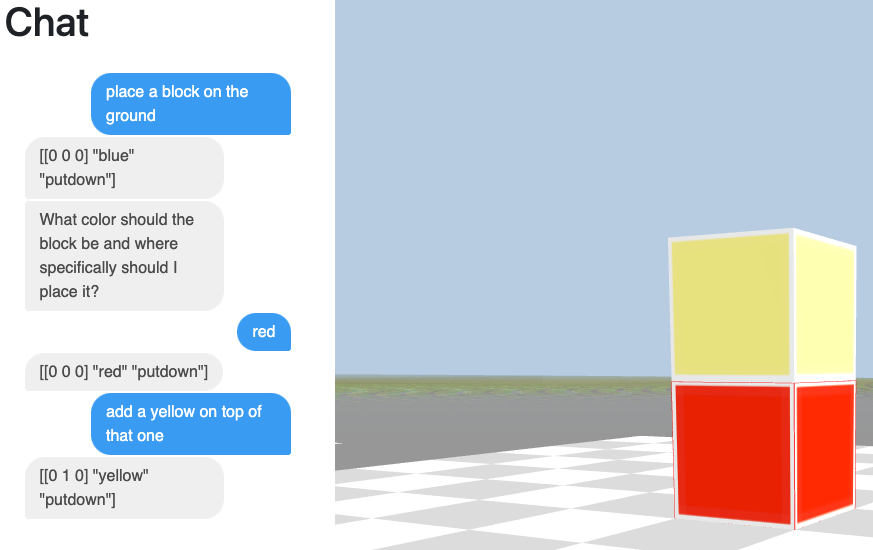}{Builder clarification and reference (GPT-4)}

In Figure \ref{fig:builder-qs-and-ref} we deliberately give an incomplete instruction, ``\textit{place a block on the ground}''. We can see GPT-4 identifies the missing properties, asking for clarification on the colour and position with the question: ``\textit{What color should the block be and where specifically should I place it?}''.  When the builder responds ``\textit{red}'' , this is resolved. We also should here a simple example of resolving reference to an existing block.

\fig[1]{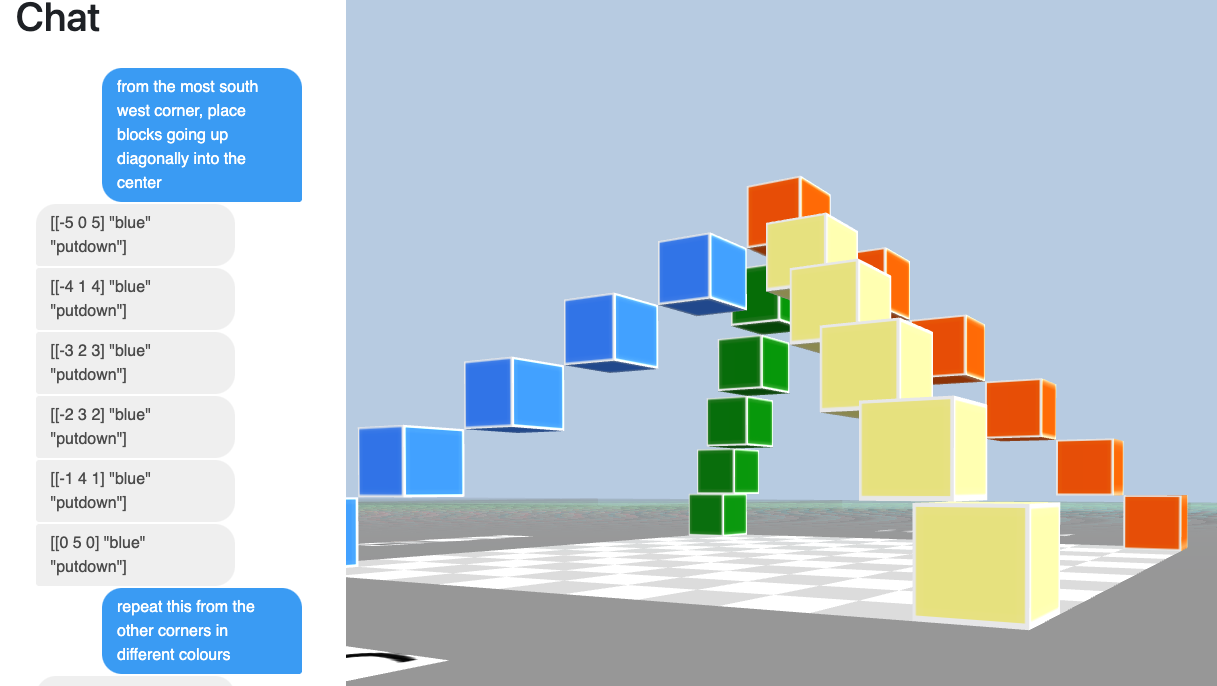}{Builder complex structures (GPT-4)}
In Figure \ref{fig:builder-diag-complex},
we can see that GPT-4 will accept descriptions of quite complex structures, references to world positions, and continue a pattern between those positions.  In addition, it will accept references that point to groups of blocks or existing structures/word spaces and carry out repetition.

\subsection{Architect}


\fig[1]{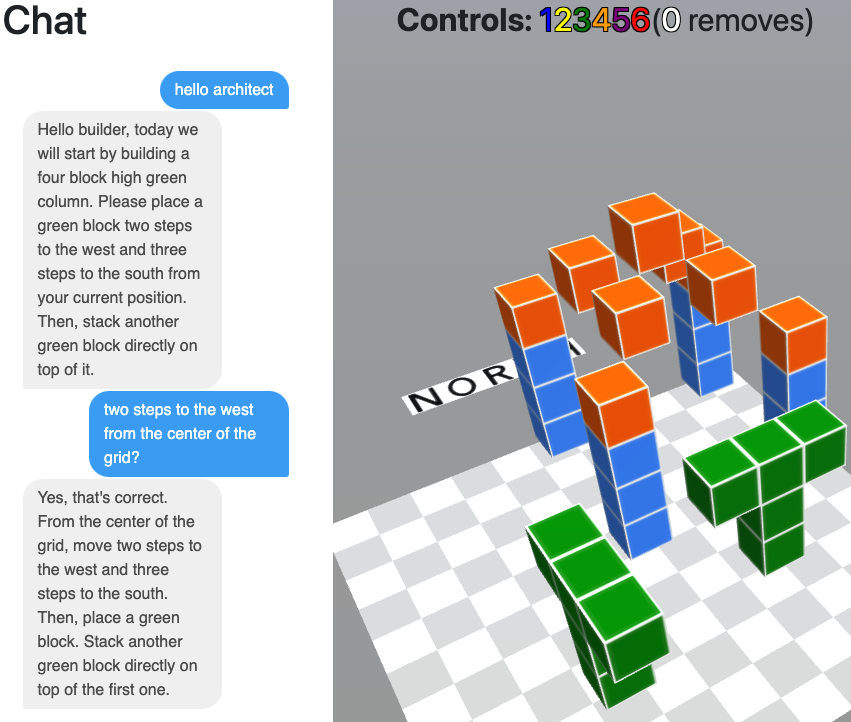}{Architect - Simple Structure (GPT-4)}
In Figure \ref{fig:architect-simple}, for the purpose of illustrating the example, the target structure is shown (this would normally be hidden from the builder). We can see, for simpler structures, the architect gives accurate instructions, can describe groups of placements (e.g. ``column''), respond to questions and provide clarification.  

\fig[1]{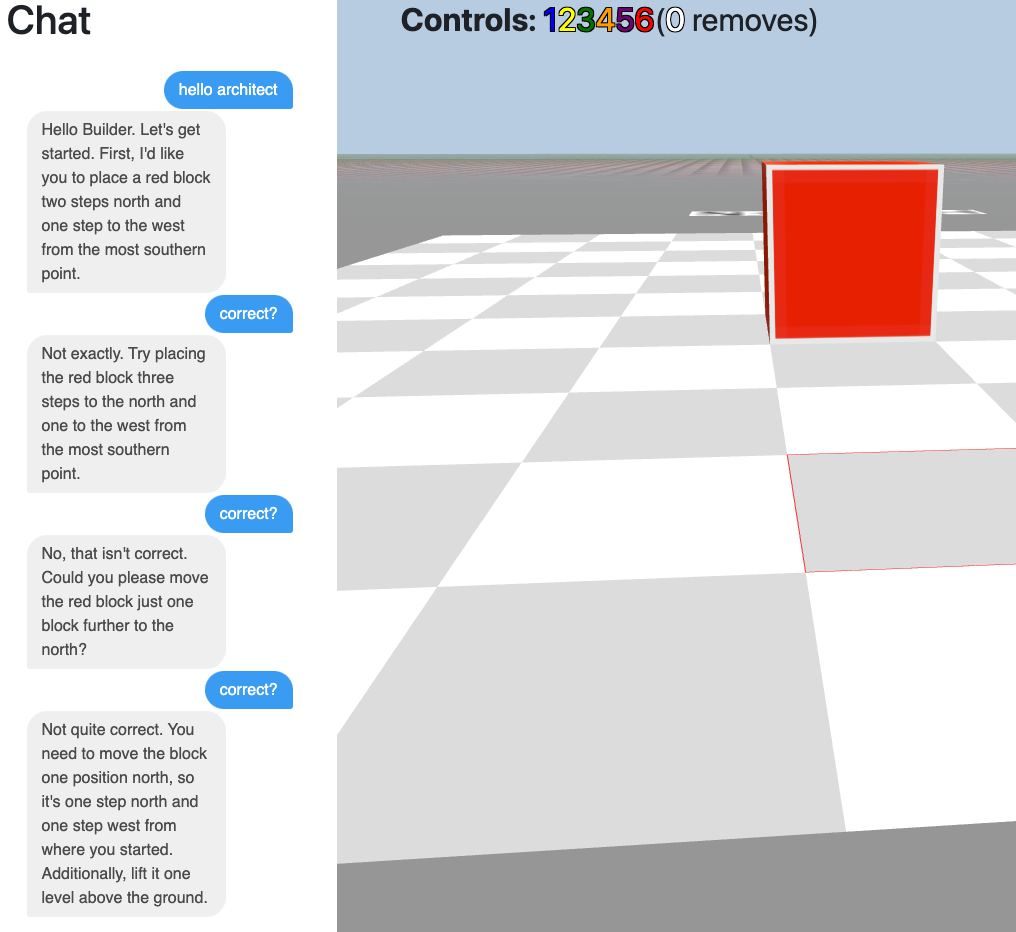}{Architect - Complex Structure (GPT-4)}
However, as the structures increase in complexity (see Figure \ref{fig:architect-complex}), the architect struggles to start. Each time the builder follows the architects instruction, it revises it getting one step closer to the true goal. The architect does eventually arrive at the appropriate instruction, after repeated exchanges between the builder and architect.


\section{Conclusion}

In this work we demonstrate LLMs acting as agents in a Minecraft-like block world task. The builder performs favourably over past bespoke models.  In addition, the LLM has an existing built in capability to ask and answer questions.  In future work, we will look at methods of improving openly available LLM models to close the gap with the fine tuned baseline, ranging from prompt engineering (e.g. introducing few-shot prompting etc.) to fine-tuning and other methods.  Our early experiments with the architect does show some challenges, and further work needs to be done to give a quantitative evaluation against existing architect models.

\section*{Acknowledgements}

This research was funded by ARCIDUCA, EPSRC EP/W001632/1
\bibliography{main}

\appendix

\section*{Appendix}
\label{sec:appendix}

\subsection*{Architect prompt}\label{sec:architectprompt}

\small{\texttt{You are an agent in a voxel world, where the most northernly point is 0,0,-5; the most westerly point -5,0,0; the most eastern point is 5,0,0; the most southern 0,0,5 and the y-axis is up and down, with y=0 being the minimum. Your task is to give instructions to a human to place blocks to achieve the target world state: <<target\_world\_state>> where the target world instructions are in the format [[x,y,z,color],...].  Give easy to interpret instructions, do not directly mention the coordinates. The builder will respond with the coordinates of the blocks they have placed in the same format.  Don't ask for coordinates, they will always be given. Avoid long instructions with multiple steps and start building the structure from the ground up.}}

\subsection*{Builder prompt}\label{sec:builderprompt}
\small{\texttt{You are an agent in a voxel world, where the most northernly point is 0,0,-5; the most westerly point -5,0,0; the most eastern point is 5,0,0; the most southern 0,0,5 and the y-axis is up and down, with y=0 being the minimum. Describe the coordinates of the blocks their colours (must be one of: blue, yellow, green, orange, purple, red) and whether the action is to add or remove them, your confidence in your interpretation of the instruction and optionally a question if the instruction is potentially unclear, in the JSON format: \{"add": [[x,y,z,color], ...], "remove": [[x,y,z,color], ...], "confidence": 0.0, "question": "..."\}. Give the JSON only, no additional dialog.}}

\subsection{Transcripts}

\paragraph{Transcript for Figure \ref{fig:builder-qs-and-ref}:}
\begin{description}
\itemsep0em
\item[Architect] \textit{Place a stone on the ground}
\item[Builder] \texttt{[[0 0 0]] "blue" "putdown"]}
\item[Builder] \textit{What colour should the block be and where specifically should I place it?}
\item[Architect] \textit{red}
\item[Builder] \texttt{[[0 0 0]] "red" "putdown"]}
\item[Architect] \textit{add a yellow on top of that one}
\item[Builder] \texttt{[[0 1 0] "yellow" "putdown"]}
\end{description}

\paragraph{Transcript for Figure \ref{fig:builder-diag-complex}:}
\begin{description}
\itemsep0em
\item[Architect] \textit{from the most south west corner, place blocks going up diagonally into the center}
\item[Builder] \texttt{[[-5 0 5]] "blue" "putdown"],[[-4 1 4] "blue" "putdown"],[[-3 2 3] "blue" "putdown"],[[-2 3 2] "blue" "putdown"],[[-1 4 1] "blue" "putdown"],[[0 5 0] "blue" "putdown"]}
\item[Architect] \textit{repeat this from the other corners in different colours}
\end{description}

\paragraph{Transcript for Figure \ref{fig:architect-simple}:}
\begin{description}
\itemsep0em
\item[Builder] \textit{hello architect}
\item[Architect] \textit{Hello builder, today we will start by building a four block high green column. Please place a green block two steps to the west and three steps to the south from your current position. Then, stack another green block directly on top of it.}
\item[Builder] \textit{two steps to the west from the center of the grid?}
\item[Architect] \textit{Yes, that's correct. From the center of the grid move two steps to the west and three steps to the south. Then, place a green block. Stack another green block directly on top of the first one.}
\end{description}

\paragraph{Transcript for Figure \ref{fig:architect-complex}:}
\begin{description}
\itemsep0em
\item[Builder] \textit{hello architect}
\item[Architect] \textit{Hello Builder. Let's get started. First, I'd like you to place a red block two steps north and one step to the west from the most southern point.}
\item[Builder] \textit{correct?}
\item[Architect] \textit{Not exactly. Try placing the red block three steps to the north and one to the west from the most southern point.}
\item[Builder] \textit{correct?}
\item[Architect] \textit{No, that isn't correct. Could you please move the red block just one block further to the north?}
\item[Builder] \textit{correct?}
\item[Architect] \textit{Not quite correct. You need to move the block one position north, so it's one step north and one step west from where you started. Additionally, lift it one level above the ground.}
\end{description}
\end{document}